\documentclass[letterpaper,10pt,journal,twoside]{IEEEtran}  %

\usepackage{amsmath} %
\usepackage{amssymb}  %
\usepackage{multirow}
\usepackage{graphicx}
\usepackage{xcolor}
\usepackage{colortbl}
\usepackage{booktabs}
\usepackage{url}

\newif\ifshowrevisioncolors
\showrevisioncolorsfalse 
\ifshowrevisioncolors
\else
  \colorlet{red}{black}
  \colorlet{blue}{black}
\fi

\makeatletter
\def\section{\@startsection{section}{1}{\z@}%
{1.8ex plus 1.5ex minus 1.5ex}%
{0.7ex plus 1ex minus 0ex}%
{\normalfont\normalsize\centering\scshape}}
\def\subsection{\@startsection{subsection}{2}{\z@}%
{2.2ex plus 1.5ex minus 1.5ex}%
{0.7ex plus .5ex minus 0ex}%
{\normalfont\normalsize\itshape}}
\makeatother
\AddToHook{env/equation/begin}{%
  \setlength{\abovedisplayskip}{2.5pt minus 1pt}
  \setlength{\belowdisplayskip}{2.5pt minus 1pt}
  \setlength{\abovedisplayshortskip}{0pt}
  \setlength{\belowdisplayshortskip}{3pt minus 1pt}
  \setlength{\jot}{2pt}
}

\title{Failure Detection for Surgical Robot Imitation Policies
via Flow-Matching World Modeling}

\author{Zhefeng Huang$^{1}$, Yilin Cai$^{1}$,
Ankit Patel$^{2}$, Mohammad Hajiha$^{3}$,
Brendan Browne$^{3}$, and Yue Chen$^{4}$%
\thanks{Received: July 9, 2026; revised October 25, 2026;
accepted November 23, 2026.
This paper was recommended for publication by Editor
Pietro Valdastri upon evaluation of the Associate Editor
and reviewers' comments.
This work was supported by NSF Awards 2339202 and 2514173,
and in part by NIH/NIBIB Award R01EB034359 and
NIH/NHLBI Award R01HL183894.
The content is solely the responsibility of the authors
and does not necessarily represent the official views
of NSF or NIH.
(Corresponding author: Yue Chen.)}%
\thanks{\textcopyright{} 2026 IEEE.
Personal use of this material is permitted.
Permission from IEEE must be obtained for all other
uses, in any current or future media, including
reprinting/republishing this material for advertising
or promotional purposes, creating new collective
works, for resale or redistribution to servers or
lists, or reuse of any copyrighted component of
this work in other works.}%
\thanks{$^{1}$Z. Huang and Y. Cai are with the Department
of Mechanical Engineering, Georgia Institute of Technology,
Atlanta, GA, USA (e-mail: zhuang480@gatech.edu;
yilincai@gatech.edu).}%
\thanks{$^{2}$A. Patel is with the Division of General and
GI Surgery, Emory University, Atlanta, GA, USA
(e-mail: apatel7@emory.edu).}%
\thanks{$^{3}$M. Hajiha and B. Browne are with the Department
of Urology, Emory University, Atlanta, GA, USA
(e-mail: mohammad.hajiha@emory.edu;
brendan.michael.browne@emory.edu).}%
\thanks{$^{4}$Y. Chen is with the Department of Biomedical
Engineering, Georgia Institute of Technology/Emory
University, Atlanta, GA, USA
(e-mail: yue.chen@bme.gatech.edu).}%
}

\begin{document}
\maketitle
\raggedbottom 

\bstctlcite{IEEE:BSTcontrol}

\begin{abstract}

Imitation learning has shown increasing promise for autonomous robotic surgery, yet safe deployment remains challenging due to the safety-critical nature of surgical tasks and the complexity and variability of surgical environments.
Failure detection is therefore an essential safeguard, but its development remains difficult due to the challenges of scarce failure data, highly variable manipulation dynamics, and the need to balance missed detections against disruptive false alarms.
To address these challenges, we introduce FoMo-FD (Flow-Matching World Model for Failure Detection), a failure detection method that learns nominal short-horizon visual dynamics with an action-conditioned flow-matching world model. FoMo-FD scores the inverse-transport nonconformity of observed endpoint latents, enabling window-level detection of deviations from learned nominal dynamics without requiring failure demonstrations. Detection thresholds are obtained by conformal calibration on successful policy rollouts, yielding task-specific alarms without assuming future failure types. We evaluate FoMo-FD on four surgically relevant manipulation tasks with twenty failure modes across simulation and real-robot experiments using the da Vinci Research Kit (dVRK). In the wrist view, the main model achieves a \(95.6\%\) failure detection rate (FDR) at a \(2.5\%\) false alarm rate (FAR), exceeding the FDR of the evaluated observation-based, action-based, and vision-language baselines and alternative scores from the same world model. Across four training seeds, mean wrist-view FDR and FAR are \(93.4\%\) and \(4.7\%\), respectively. Online replay yields a mean onset-relative detection delay of \(0.98\)~s for wrist-view failures detected offline. Project website: \url{https://janusmaple.github.io/FoMoFD/}

\end{abstract}

\begin{IEEEkeywords}
Failure Detection; Surgical Robotics: Planning; Imitation Learning
\end{IEEEkeywords}

\newpage
\section{Introduction}

\IEEEPARstart{A}{utonomy} in robot-assisted surgery has the potential to improve procedural consistency, reduce surgeon workload, and expand access to high-quality care. Although most surgical robotic systems remain teleoperated, their use provides a practical path toward autonomy. Expert demonstrations can be collected directly during routine procedures and used as training data. Motivated by this opportunity, recent work has explored imitation learning for surgical visuomotor policies from teleoperated demonstrations \cite{kawaharazuka2024robotic, kim2025surgical, kim2025srt, haworth2026suturebot}. However, learning from demonstrations alone does not guarantee reliable execution. In visually complex and safety-critical surgical environments, deployment shifts can lead to policy failures, making execution monitoring and timely intervention essential. Existing approaches to surgical execution monitoring include gesture-aware kinematic error detection \cite{yasar2020real, li2022runtime}, visual and multimodal detection using gesture or activity context \cite{shao2024think, roodabeh2026realtime}, and confidence-based supervised human handoff \cite{kam2021confidence}. Although effective in their respective settings, these approaches train their monitoring models using labeled errors or task-specific success and failure outcomes. Other work has explored uncertainty-aware learned policies that estimate epistemic uncertainty from surgical policy ensembles and use output disagreement to identify failure-prone executions \cite{thompson2025early, empleo2026safe}. However, policy-output uncertainty provides only an indirect failure signal and may not reliably capture deviations from nominal execution.

This limitation points to a broader runtime monitoring problem for learned robot policies, in which policy confidence alone is insufficient for detecting failures during deployment. Sentinel \cite{agia2025unpacking} distinguishes between erratic failures, where a policy produces temporally inconsistent actions, and task-progression failures, where  the policy remains consistent but fails to advance the task, and prior surgical studies also distinguish executional errors from procedural errors \cite{hutchinson2022analysis}. Monitors based on vision--language models (VLMs), including Sentinel's progress monitor, Code-as-Monitor \cite{agia2025unpacking, zhou2025code}, and CARES \cite{low2025cares}, provide complementary semantic reasoning but rely on task instructions, instruction-derived visual constraints, or error-specific prompts. Additionally, SAFE \cite{gu2026safe} enables timely detection using vision-language-action (VLA) internal features, but requires failure episodes for training. In contrast, we seek a general monitoring formulation that detects low-level execution failures by characterizing nominal execution from successful episodes, without requiring failure demonstrations or task-specific descriptions.

Existing methods for timely failure detection often rely on action-level signals, such as ensemble-based policy uncertainty \cite{thompson2025early, empleo2026safe}, temporal action inconsistency \cite{agia2025unpacking}, or action-chunk entropy \cite{romer2026failure}. Yet action-only signals may miss failures caused by visual disturbances, unexpected interaction dynamics, or action-execution mismatch. In cable-driven surgical instruments, hysteresis can also cause discrepancies between commanded actions and resulting motion \cite{hwang2020efficiently}. Beyond action-only signals, visual distributional scores support detection of out-of-distribution (OOD) observations in robot failure monitors such as FAILDetect-logpZO \cite{xu2025can} and FIPER \cite{romer2026failure}, as well as image anomaly detection in CFLOW-AD \cite{gudovskiy2022cflow}. However, their visual scores are primarily computed at the observation level, assessing whether an individual frame or visual embedding appears anomalous rather than whether a short observation sequence follows nominal temporal dynamics. This distinction is important because some failures may not make any individual frame obviously OOD, but instead appear as abnormal transitions over a temporal window. RC-NF \cite{zhou2026rc} models temporal sequences of segmented object point sets using a normalizing flow conditioned on robot states and task embeddings, with conditional likelihood used for anomaly scoring. These considerations motivate a monitor that captures temporal visual deviations and uses commanded actions as additional context for interpreting the observed response. Such a monitor should operate without object segmentation or task prompts, which can be difficult to obtain or specify in surgical settings.

World models provide a way to learn the temporal structure of execution. Recent surgical world models support video prediction, data augmentation, policy learning, and evaluation \cite{lin2024world, koju2025surgical, he2025cosmoshsurgical, turkcan2025suturing, zbinden2026cosmos}, although causal and procedural plausibility can remain challenging beyond visual realism \cite{chen2025how}. Related work forecasts visual representations in pretrained feature spaces \cite{zhou2025dino, baldassarre2025features, karypidis2026dino, boduljak2025vfmf}. Beyond forecasting, world models have also been used for failure detection. Ward et al. trained a probabilistic, history-informed world model in a pretrained visual latent space and used its predictive uncertainty as a nonconformity score for conformal failure detection \cite{ward2026foundational}. While effective, uncertainty-based scoring measures how confident the model is in its forecast, rather than whether the realized execution matches learned nominal dynamics.

\begin{figure*}[!t]
    \centering
    \includegraphics[width=0.96\textwidth]{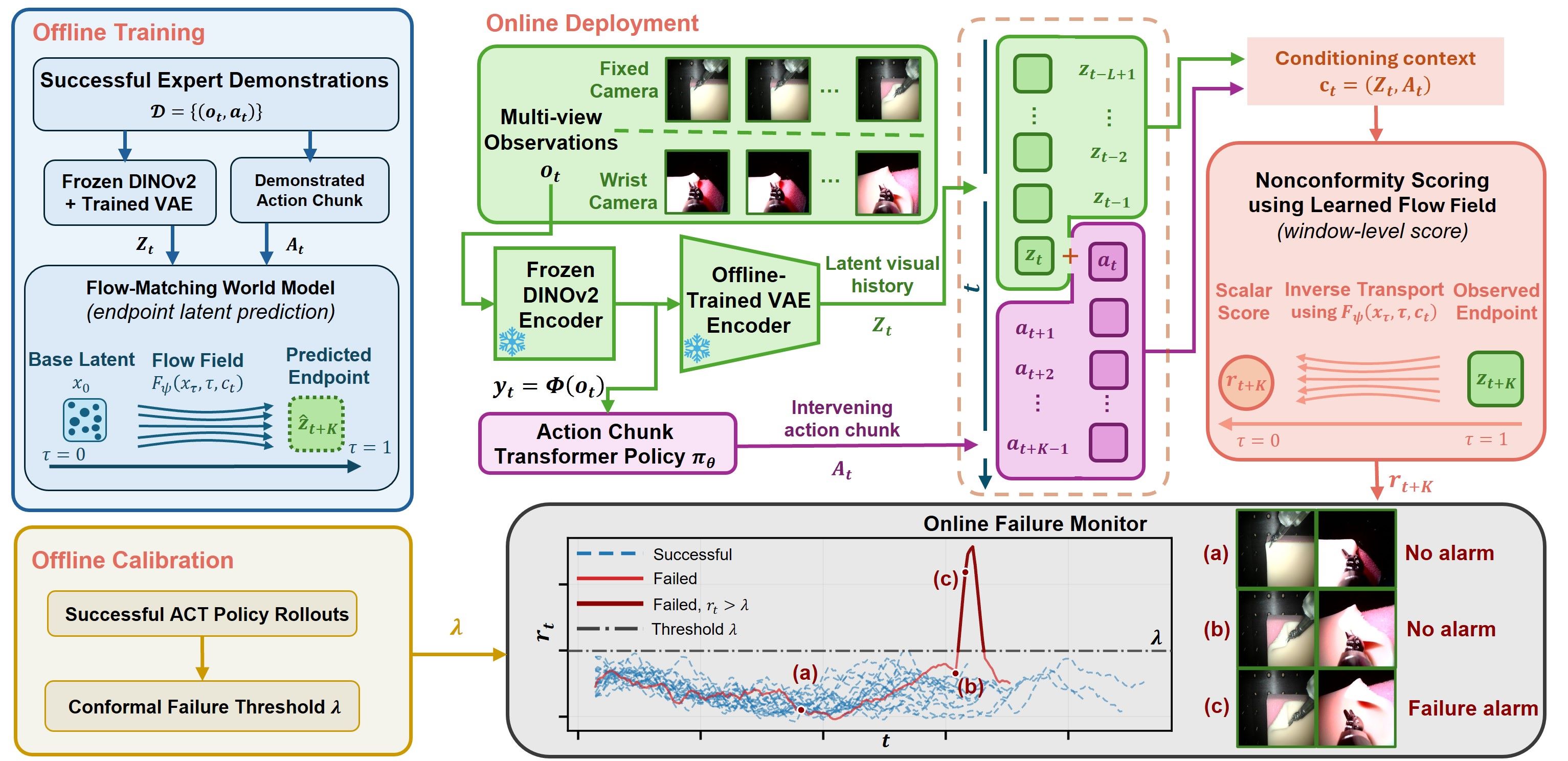}
    \caption{Overview of FoMo-FD. A frozen DINOv2 backbone and offline-trained VAEs encode observations into compact, view-specific latents. A flow-matching world model learns endpoint prediction from successful expert demonstrations, conditioned on visual history and intervening commanded actions. Successful policy rollouts provide conformal thresholds for each view. During online monitoring, the observed endpoint is scored by inverse transport under learned nominal dynamics, and each view independently raises an alarm when its nonconformity score exceeds its calibrated threshold.}
    \label{fig:overview}
\end{figure*}

To assess observed execution directly, we introduce FoMo-FD (Flow-Matching World Model for Failure Detection), a failure detection method based on a flow-matching visual world model, as illustrated in Fig.~\ref{fig:overview}. Rather than recursively predicting the next frame, FoMo-FD predicts the endpoint latent of a short observation window conditioned on the preceding observation history and the intervening robot actions. It then scores the inverse-transport nonconformity of the observed endpoint latent under the learned nominal dynamics. The world model is trained only on successful demonstrations. Drawing on conformal prediction \cite{angelopoulos2023conformal} and its applications to robotic runtime monitoring \cite{luo2024sample,lindemann2023conformal}, we calibrate detection thresholds using a small set of successful policy rollouts, enabling task-specific thresholding without failure data. We validate FoMo-FD in both simulation and real-robot da Vinci Research Kit (dVRK) experiments \cite{kazanzides2014open}, highlighting how action-conditioned, window-level latent dynamics can reveal execution inconsistencies during surgical manipulation.

In summary, this work makes three contributions:
\begin{itemize}
\item We develop an endpoint-prediction flow-matching world model for surgical visual dynamics, which predicts the latent endpoint of a short execution window conditioned on visual history and intended robot actions.
\item We formulate failure detection as window-level execution monitoring by scoring the inverse-transport nonconformity of observed endpoint latents and calibrating thresholds from successful rollouts, without requiring failure demonstrations.
\item We evaluate the framework on four surgery-inspired manipulation tasks across simulation and real-robot dVRK experiments with twenty diverse failure modes, demonstrating higher wrist-view FDR than the evaluated baselines while maintaining low FAR.
\end{itemize}

\section{Problem Formulation}

We consider a visuomotor imitation policy deployed in a partially observable robotic manipulation environment. At every timestep \(t\), the robot receives a new observation \(o_t\) and the policy \(\pi_\theta\) outputs a commanded action \(a_t = \pi_\theta(h_t)\) based on the available observation-action history \(h_t = (o_{1:t}, a_{1:t-1})\). The policy is trained on an expert demonstration dataset \(\mathcal{D}\) to approximate the expert action mapping conditioned on observation-action history.

The goal is to learn a runtime failure detector that monitors execution and raises an alarm when behavior deviates from nominal successful execution. For a window of length \(H\) ending at timestep \(t\), let \(w_t = (o_{t-H+1:t}, a_{t-H+1:t-1})\) denote the recent observation-action history. A detector computes a nonconformity score \(r_t = f(w_t)\), where larger values indicate greater deviation from nominal execution. Given a threshold \(\lambda\) calibrated from successful rollouts, the detector induces a binary classifier \(\mathcal{C}\):
\begin{equation}
\mathcal{C}(w_t) =
\begin{cases}
1, & f(w_t) > \lambda, \quad \text{abnormal}\\
0, & f(w_t) \leq \lambda, \quad \text{nominal},
\end{cases}
\label{eq:failure_classifier}
\end{equation}
\noindent where \(\mathcal{C}(w_t)=1\) indicates a detected failure.

\section{Method}

\subsection{World Modeling in Latent Vision Space}

We model nominal short-horizon visual dynamics in a compact latent space. A frozen DINOv2 encoder \cite{oquab2024dinov2} extracts patch features from each observation, which we compress with an offline-trained variational autoencoder (VAE). We adapt VFMF's flow-matching framework \cite{boduljak2025vfmf} for failure detection by conditioning on the intervening robot actions and predicting the endpoint latent of a short execution window instead of autoregressively predicting future latents. The overall pipeline is summarized in Fig.~\ref{fig:overview}.

Formally, let \(\Phi\) denote the frozen DINOv2 encoder, and let \(y_t^v=\Phi(o_t^v)\) denote the DINOv2 patch features extracted from observation \(o_t^v\) in camera view \(v\), with \(v \in \{\mathrm{fix}, \mathrm{wrist}\}\). For each view, we train a feature-space \(\beta\)-VAE with encoder \(E_\phi^v\) and reconstruction decoder \(R_\eta^v\). Given \(y_t^v\), the encoder outputs \((\mu_t^v,\sigma_t^v)=E_\phi^v(y_t^v)\), which parameterize a diagonal Gaussian distribution denoted by \(\mathcal{N}_{E_\phi^v(y_t^v)}\). The resulting latent tensor \(z_t^v \in \mathbb{R}^{N_z^h \times N_z^w \times C_z}\), where \(N_z^h\), \(N_z^w\), and \(C_z\) denote the latent height, width, and channel dimensions, respectively, is sampled from this distribution as \(z_t^v=\mu_t^v+\sigma_t^v\odot\epsilon\), with \(\epsilon\sim\mathcal{N}(0,I)\). The VAE is trained with
\begingroup
\predisplaypenalty=0
\clubpenalty=10000
\begin{equation}
\begin{aligned}
\mathcal{L}_{\mathrm{VAE}}
=&
\sum\nolimits_v\mathbb{E}_{y_t^v,\,z_t^v}\left[\left\|R_\eta^v(z_t^v)-y_t^v\right\|_2^2\right]\\
&+
\beta\sum\nolimits_v\mathbb{E}_{y_t^v}\left[D_{\mathrm{KL}}\left(\mathcal{N}_{E_\phi^v(y_t^v)}\;\|\;\mathcal{N}(0,I)\right)\right]\\
&+
\gamma\sum\nolimits_v\mathbb{E}_{y_t^v}\left[\mathcal{L}_{\mathrm{util}}^v\right],
\end{aligned}
\end{equation}
where \(\mathcal{L}_{\mathrm{util}}^v\) is a channel-utilization regularizer computed from the normalized per-channel variance of the encoder mean \(\mu_t^v\). This regularizer discourages information from concentrating in only a few latent channels. After training, the VAE is frozen, and we use the deterministic posterior mode \(z_t^v=\mu_t^v\) as the compact latent representation for all subsequent world-model training and scoring.
\par\endgroup

Given the encoded latent sequence, we define latent history length \(L\) and prediction horizon \(K\). For each view \(v\), let
\begin{equation}
Z_t^v = z_{t-L+1:t}^v, \qquad A_t = a_{t:t+K-1}\textcolor{red}{,}
\end{equation}
where \(Z_t^v\) denotes the latent visual history and \(A_t\) denotes the commanded action chunk over the prediction horizon. Together, they form the conditioning context \(c_t^v=(Z_t^v,A_t)\). For each view, the world model learns a parameterized approximation \(p_\psi^v(z_{t+K}^v \mid c_t^v)\) to the nominal endpoint distribution from successful executions.

We instantiate this distribution with flow matching \cite{lipman2023flow}. For each enabled view, set \(x_1^v=z_{t+K}^v\) and sample \(x_0^v \sim \mathcal{N}(0,\sigma^2 I)\) from the base distribution. Along the standard linear path, for \(\tau \sim \mathcal{U}(0,1)\),
\begin{equation}
x_\tau^v = (1-\tau)x_0^v + \tau x_1^v, \qquad u_\tau^v = x_1^v - x_0^v\textcolor{red}{.}
\end{equation}

The model learns a conditional vector field \(F_\psi^v(x_\tau^v,\tau,c_t^v)\) to match the target velocity \(u_\tau^v\) by minimizing:
\begin{equation}
\mathcal{L}_{\mathrm{FM}}=\sum_v\mathbb{E}_{x_0^v,x_1^v,\tau}\left[\left\|F_\psi^v(x_\tau^v,\tau,c^v_t)-u_\tau^v\right\|_2^2\right]\textcolor{red}{.}
\end{equation}

For the fixed- and wrist-camera inputs in this work, the objective sums the losses from two view-specific branches. Each branch receives the latent history \(Z_t^v\) and the interpolated endpoint state \(x_\tau^v\), projects them to a hidden dimension, and processes them with a transformer backbone using temporal attention over the history dimension and spatial attention over the latent tensor. The visual branches are view-specific, while the action conditioner is shared across views. The same action chunk \(A_t\) provides action tokens for cross-attention and a global action embedding for adaptive modulation. The conditioned transformer backbone then feeds a final layer that outputs the velocity \(F_\psi^v(x_\tau^v,\tau,c_t^v)\), which transports \(x_\tau^v\) along the path toward the endpoint latent \(z_{t+K}^v\).

In addition to the endpoint flow-matching objective, we add an auxiliary one-step latent prediction loss to regularize the visual-action features learned by the world-model backbone. For each view, an auxiliary head \(G_\psi^v(c_t^v)\) predicts the next latent \(\hat{z}_{t+1}^v\) from the same conditioning context and is trained with
\begin{equation}
\mathcal{L}_{\mathrm{aux}}=\sum_v\mathbb{E}\left[\left\|G_\psi^v(c_t^v) - z_{t+1}^v\right\|_2^2\right].
\end{equation}
The full training objective is
\begin{equation}
\mathcal{L}=\lambda_{\mathrm{FM}}\mathcal{L}_{\mathrm{FM}}+\lambda_{\mathrm{aux}}\mathcal{L}_{\mathrm{aux}}.
\end{equation}
The auxiliary head is used only during training and is not used for nonconformity scoring at test time.

\subsection{Nonconformity Scoring and Conformal Failure Detection}

At test time, FoMo-FD scores whether the observed endpoint latent is consistent with the preceding visual history and commanded action. For a scoring window ending at time \(t\), the current observation \(o_t\) is first encoded into visual latents \(z_t^v\). The nonconformity score \(r_t\) is computed under the learned transport conditioned on \(c_{t-K}\), which contains the \(L\)-step latent history ending at \(t-K\) and the subsequent \(K\)-step action chunk. Thus, the scoring window length in Eq.~\eqref{eq:failure_classifier} corresponds to \(H=L+K\). Scoring is performed through inverse transport without requiring an explicit sampled future prediction. Following a log-\(p(z_0)\)-style inverse-flow criterion \cite{xu2025can}, we set \(x_1^v=z_t^v\) and integrate the learned flow-matching ordinary differential equation (ODE) backward:
\begin{equation}
\frac{dx_\tau^v}{d\tau} = F_\psi^v(x_\tau^v,\tau,c_{t-K}^v), \quad r_t^v = \left\| \frac{\hat{\epsilon}_t^v}{\sigma} \right\|_2^2,
\end{equation}
where \(\hat{\epsilon}_t^v=x_{\tau=0}^v\) is the inverse-transported endpoint. Its squared normalized distance from the Gaussian base mean serves as a measure of deviation from nominal dynamics. In our implementation, scores are computed separately for each view and monitored independently.

To obtain failure thresholds without failure data, we collect \(N\) successful calibration rollouts. For each view \(v\) and target miscoverage level \(\alpha\in[1/(N+1),1)\), we compute the episode-level peak scores \(s_j^v=\max_t r_{t,j}^v\) and sort them as \(s_{(1)}^v\leq\cdots\leq s_{(N)}^v\). The conformal threshold is then set to
\begin{equation}
\lambda^v = s_{\left(\left\lceil (N+1)(1-\alpha)\right\rceil\right)}^v.
\end{equation}

During evaluation, view \(v\) raises an alarm whenever \(r_t^v>\lambda^v\). Under exchangeability of successful calibration and test rollouts, this threshold bounds the marginal probability of any alarm in a new successful rollout by \(\alpha\) for each view. Because calibration uses only successful executions, the resulting threshold is task-specific while avoiding assumptions about the type or timing of future failures.

\section{Experiments}

We evaluate FoMo-FD on four surgically relevant manipulation tasks spanning both simulation and real-robot dVRK experiments. The simulated tasks are needle pickup and ring over post, while the real-robot tasks are tissue retraction and shunt insertion. For each task, the detector is calibrated using successful rollouts and evaluated on held-out successful executions and task-specific failure modes.

\subsection{Experimental Setup}

\begin{figure}[!t]
    \centering
    \includegraphics[width=0.48\textwidth]{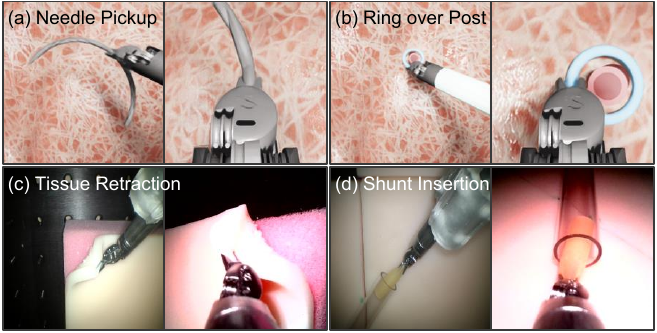}
    \caption{Representative observations from the four tasks. Each panel shows the fixed-camera view on the left and the wrist-camera view on the right. Top row: simulated needle pickup and ring over post. Bottom row: real-robot tissue retraction and shunt insertion.}
    \label{fig:tasks}
\end{figure}

\noindent\textbf{Tasks and platforms.}
The needle-pickup and ring-over-post experiments are conducted in NVIDIA Isaac Sim using anatomically realistic organ scenes adapted from the SuFIA-BC environment~\cite{moghani2025sufia}. During demonstration collection, a physical dVRK master tool manipulator (MTM) is used to teleoperate the simulated patient-side manipulator (PSM). The tissue-retraction and shunt-insertion experiments are conducted on a physical dVRK. Each setup provides a fixed-camera view of the workspace and a wrist-camera view of the local tool-object interaction, and representative observations from the four tasks are depicted in Fig.~\ref{fig:tasks}. \textcolor{red}{
For simulated tasks, observations and policy actions
are updated at 30 Hz in simulation time (approximately
6 Hz in wall time). For real-robot tasks, both are
updated at approximately 10 Hz, with robot commands
interpolated and published at 100 Hz.
}

\noindent\textbf{Policy and detector training.}
For each task, we train a separate Action Chunking Transformer (ACT) policy \cite{zhao2023learning} from teleoperated expert demonstrations. One operator collected 250 expert episodes per simulated task and 100 per real-robot task. ACT uses all of these for training. The policy and detector use the same frozen DINOv2 ViT-B/14 backbone with registers on \(224\times224\) images. For the detector, we first train view-specific feature-space VAEs on the backbone's patch features from the same expert demonstrations, using \(\beta=0.01\) and channel-utilization weight \(\gamma=0.5\). After training, the VAEs are frozen and their deterministic posterior modes are used to encode these features into latent tensors of size \(N_z^h=N_z^w=16\) with \(C_z=8\) channels. The flow-matching world model is then trained on the same expert demonstrations in this compressed latent space. The visual VAEs and world model share the same episode-level training/validation split: 225/25 episodes per simulated task and 90/10 per real-robot task. For the main model, we use a latent history length of \(L=4\), consistent with VFMF's maximum context length~\cite{boduljak2025vfmf}, and an endpoint prediction horizon of \(K=4\), matching ACT's predicted action-chunk length, yielding a total monitored window length of \(H=L+K=8\). The world model is trained with \(\lambda_{\mathrm{FM}}=1.0\), \(\lambda_{\mathrm{aux}}=0.5\), and base standard deviation \(\sigma=1\); scoring uses one fourth-order Runge--Kutta (RK4) step per ODE solve (four model evaluations). Policy rollouts are used only for detector calibration and evaluation, with the fixed and wrist views scored independently using separate nonconformity scores and conformal thresholds.

\begin{table*}[!t]
\begingroup
\color{black}
\arrayrulecolor{black}
\fontsize{8}{9.5}\selectfont
\makeatletter
\renewcommand{\fnum@table}{\textcolor{black}{TABLE~\thetable}}
\makeatother
\caption{\textcolor{black}{Failure detection across four tasks}}
\label{tab:failure_detection}
\centering
\fontsize{8}{9.5}\selectfont
\setlength{\tabcolsep}{1.1pt}
\renewcommand{\arraystretch}{1.00}
\newcommand{\TableICount}[2]{\makebox[1em][r]{#1}/#2}
\newsavebox{\TableIBox}
\newcommand{\TableIContent}{%
\toprule
& & \multicolumn{8}{c}{\textbf{WM scores}}
& \multicolumn{6}{c}{\textbf{Observation}}
& \multicolumn{2}{c}{\textbf{Action}} \\
\cmidrule(lr){3-10}\cmidrule(lr){11-16}\cmidrule(l){17-18}
& & \multicolumn{2}{c}{\textbf{WM-NC} (ours)}
& \multicolumn{2}{c}{\textbf{NoAct}}
& \multicolumn{2}{c}{\textbf{WM-PE}}
& \multicolumn{2}{c}{\textbf{WM-PU}}
& \multicolumn{2}{c}{\textbf{logpZO}}
& \multicolumn{2}{c}{\textbf{RND}}
& \multicolumn{2}{c}{\textbf{VLM}}
& \textbf{ACE} & \textbf{AEU} \\
\cmidrule(lr){3-4}\cmidrule(lr){5-6}\cmidrule(lr){7-8}\cmidrule(lr){9-10}
\cmidrule(lr){11-12}\cmidrule(lr){13-14}\cmidrule(lr){15-16}\cmidrule(l){17-18}
\textbf{Failure mode} & \textbf{Cat.}
& Fixed & Wrist & Fixed & Wrist & Fixed & Wrist & Fixed & Wrist
& Fixed & Wrist & Fixed & Wrist & Fixed & Wrist & Both & Both \\
\midrule
\multicolumn{18}{@{}l@{}}{\textbf{(A) Calibrated constant thresholds: per-mode counts and pooled rates}} \\
& & \multicolumn{16}{c}{\textbf{Needle Pickup (Sim.)} --- ACT natural success: 10/20 (50\%)} \\
\rowcolor{black!9}
No failure & -- & \textbf{\TableICount{0}{20}} & \textbf{\TableICount{0}{20}} & \TableICount{1}{20} & \TableICount{1}{20} & \TableICount{1}{20} & \textbf{\TableICount{0}{20}} & \textbf{\TableICount{0}{20}} & \textbf{\TableICount{0}{20}} & \TableICount{2}{20} & \TableICount{1}{20} & \textbf{\TableICount{0}{20}} & \textbf{\TableICount{0}{20}} & \TableICount{5}{20} & \TableICount{0}{20} & \textbf{\TableICount{0}{20}} & \textbf{\TableICount{0}{20}} \\
\rowcolor{black!9}
Needle Drop & I & \TableICount{12}{20} & \textbf{\TableICount{20}{20}} & \TableICount{17}{20} & \TableICount{17}{20} & \TableICount{13}{20} & \TableICount{3}{20} & \TableICount{2}{20} & \TableICount{1}{20} & \TableICount{15}{20} & \TableICount{12}{20} & \TableICount{5}{20} & \TableICount{5}{20} & \TableICount{1}{20} & \TableICount{0}{20} & \TableICount{19}{20} & \TableICount{19}{20} \\
\rowcolor{black!9}
Insecure Grasp & I & \TableICount{0}{20} & \TableICount{18}{20} & \TableICount{14}{20} & \TableICount{13}{20} & \TableICount{1}{20} & \TableICount{5}{20} & \TableICount{0}{20} & \TableICount{2}{20} & \TableICount{15}{20} & \TableICount{5}{20} & \TableICount{5}{20} & \TableICount{1}{20} & \TableICount{0}{20} & \TableICount{0}{20} & \textbf{\TableICount{20}{20}} & \TableICount{17}{20} \\
Blood Occlusion & IV & \textbf{\TableICount{20}{20}} & \textbf{\TableICount{20}{20}} & \textbf{\TableICount{20}{20}} & \textbf{\TableICount{20}{20}} & \TableICount{7}{20} & \TableICount{1}{20} & \TableICount{0}{20} & \TableICount{1}{20} & \textbf{\TableICount{20}{20}} & \textbf{\TableICount{20}{20}} & \TableICount{13}{20} & \TableICount{14}{20} & \TableICount{9}{20} & \TableICount{1}{20} & \TableICount{3}{20} & \TableICount{0}{20} \\
OOD Initial Condition & II & \TableICount{3}{20} & \textbf{\TableICount{20}{20}} & \TableICount{10}{20} & \textbf{\TableICount{20}{20}} & \TableICount{3}{20} & \TableICount{6}{20} & \TableICount{0}{20} & \TableICount{6}{20} & \TableICount{19}{20} & \textbf{\TableICount{20}{20}} & \TableICount{2}{20} & \TableICount{15}{20} & \TableICount{9}{20} & \TableICount{0}{20} & \TableICount{11}{20} & \TableICount{2}{20} \\
Execution Mismatch & IV & \TableICount{2}{20} & \textbf{\TableICount{20}{20}} & \TableICount{2}{20} & \TableICount{11}{20} & \TableICount{0}{20} & \TableICount{6}{20} & \TableICount{0}{20} & \TableICount{1}{20} & \TableICount{4}{20} & \TableICount{6}{20} & \TableICount{0}{20} & \TableICount{1}{20} & \TableICount{1}{20} & \TableICount{0}{20} & \TableICount{6}{20} & \TableICount{3}{20} \\
Needle Slippage & III & \TableICount{15}{20} & \textbf{\TableICount{20}{20}} & \TableICount{18}{20} & \TableICount{19}{20} & \TableICount{10}{20} & \TableICount{0}{20} & \TableICount{0}{20} & \TableICount{0}{20} & \TableICount{5}{20} & \TableICount{4}{20} & \TableICount{0}{20} & \TableICount{0}{20} & \TableICount{3}{20} & \TableICount{0}{20} & \TableICount{0}{20} & \TableICount{0}{20} \\
\midrule
& & \multicolumn{16}{c}{\textbf{Ring Over Post (Sim.)} --- ACT natural success: 15/20 (75\%)} \\
\rowcolor{black!9}
No failure & -- & \textbf{\TableICount{0}{20}} & \TableICount{2}{20} & \textbf{\TableICount{0}{20}} & \TableICount{1}{20} & \textbf{\TableICount{0}{20}} & \textbf{\TableICount{0}{20}} & \textbf{\TableICount{0}{20}} & \textbf{\TableICount{0}{20}} & \textbf{\TableICount{0}{20}} & \textbf{\TableICount{0}{20}} & \TableICount{1}{20} & \TableICount{1}{20} & \TableICount{5}{20} & \TableICount{13}{20} & \textbf{\TableICount{0}{20}} & \TableICount{1}{20} \\
\rowcolor{black!9}
Post Misalignment & II & \TableICount{0}{20} & \TableICount{13}{20} & \TableICount{0}{20} & \textbf{\TableICount{14}{20}} & \TableICount{0}{20} & \TableICount{1}{20} & \TableICount{0}{20} & \TableICount{0}{20} & \TableICount{1}{20} & \TableICount{13}{20} & \TableICount{2}{20} & \TableICount{3}{20} & \TableICount{0}{20} & \TableICount{15}{20} & \TableICount{0}{20} & \TableICount{1}{20} \\
\rowcolor{black!9}
Insecure Grasp & I & \TableICount{0}{20} & \textbf{\TableICount{20}{20}} & \TableICount{0}{20} & \textbf{\TableICount{20}{20}} & \TableICount{0}{20} & \TableICount{3}{20} & \TableICount{0}{20} & \TableICount{0}{20} & \TableICount{1}{20} & \TableICount{12}{20} & \TableICount{0}{20} & \TableICount{12}{20} & \TableICount{5}{20} & \TableICount{3}{20} & \TableICount{4}{20} & \TableICount{11}{20} \\
Ring Drop & I & \TableICount{0}{20} & \textbf{\TableICount{19}{20}} & \TableICount{0}{20} & \TableICount{13}{20} & \TableICount{1}{20} & \TableICount{0}{20} & \TableICount{1}{20} & \TableICount{0}{20} & \TableICount{0}{20} & \TableICount{14}{20} & \TableICount{0}{20} & \TableICount{2}{20} & \TableICount{0}{20} & \TableICount{4}{20} & \TableICount{8}{20} & \TableICount{15}{20} \\
Blood Occlusion & IV & \textbf{\TableICount{20}{20}} & \textbf{\TableICount{20}{20}} & \textbf{\TableICount{20}{20}} & \textbf{\TableICount{20}{20}} & \textbf{\TableICount{20}{20}} & \TableICount{1}{20} & \TableICount{1}{20} & \TableICount{0}{20} & \textbf{\TableICount{20}{20}} & \textbf{\TableICount{20}{20}} & \TableICount{9}{20} & \TableICount{14}{20} & \TableICount{7}{20} & \TableICount{14}{20} & \TableICount{3}{20} & \TableICount{3}{20} \\
Lateral Grasp Drift & IV & \TableICount{1}{20} & \textbf{\TableICount{20}{20}} & \TableICount{1}{20} & \textbf{\TableICount{20}{20}} & \TableICount{7}{20} & \TableICount{0}{20} & \TableICount{0}{20} & \TableICount{0}{20} & \TableICount{0}{20} & \TableICount{0}{20} & \TableICount{0}{20} & \TableICount{0}{20} & \TableICount{3}{20} & \TableICount{1}{20} & \TableICount{0}{20} & \TableICount{0}{20} \\
Ring Slippage & III & \TableICount{12}{20} & \textbf{\TableICount{20}{20}} & \TableICount{0}{20} & \TableICount{13}{20} & \TableICount{0}{20} & \TableICount{0}{20} & \TableICount{0}{20} & \TableICount{0}{20} & \TableICount{0}{20} & \TableICount{2}{20} & \TableICount{2}{20} & \TableICount{1}{20} & \TableICount{14}{20} & \TableICount{9}{20} & \TableICount{0}{20} & \TableICount{0}{20} \\
\midrule
& & \multicolumn{16}{c}{\textbf{Tissue Retraction (Real)} --- ACT natural success: 20/20 (100\%)} \\
\rowcolor{black!9}
No failure & -- & \textbf{\TableICount{0}{20}} & \textbf{\TableICount{0}{20}} & \textbf{\TableICount{0}{20}} & \textbf{\TableICount{0}{20}} & \textbf{\TableICount{0}{20}} & \TableICount{4}{20} & \textbf{\TableICount{0}{20}} & \textbf{\TableICount{0}{20}} & \textbf{\TableICount{0}{20}} & \textbf{\TableICount{0}{20}} & \textbf{\TableICount{0}{20}} & \TableICount{1}{20} & \TableICount{0}{20} & \TableICount{0}{20} & \textbf{\TableICount{0}{20}} & \TableICount{1}{20} \\
Tissue Grasp Loss & I & \textbf{\TableICount{10}{10}} & \textbf{\TableICount{10}{10}} & \TableICount{8}{10} & \textbf{\TableICount{10}{10}} & \TableICount{0}{10} & \textbf{\TableICount{10}{10}} & \TableICount{0}{10} & \TableICount{0}{10} & \TableICount{5}{10} & \textbf{\TableICount{10}{10}} & \TableICount{9}{10} & \textbf{\TableICount{10}{10}} & \TableICount{1}{10} & \TableICount{0}{10} & \TableICount{0}{10} & \TableICount{6}{10} \\
Fixed Tissue Target & III & \textbf{\TableICount{10}{10}} & \textbf{\TableICount{10}{10}} & \TableICount{9}{10} & \textbf{\TableICount{10}{10}} & \TableICount{4}{10} & \textbf{\TableICount{10}{10}} & \TableICount{0}{10} & \TableICount{1}{10} & \TableICount{7}{10} & \textbf{\TableICount{10}{10}} & \TableICount{8}{10} & \textbf{\TableICount{10}{10}} & \TableICount{0}{10} & \TableICount{0}{10} & \TableICount{0}{10} & \TableICount{2}{10} \\
Tissue Disturbance & III & \textbf{\TableICount{10}{10}} & \TableICount{9}{10} & \textbf{\TableICount{10}{10}} & \textbf{\TableICount{10}{10}} & \TableICount{6}{10} & \TableICount{4}{10} & \TableICount{0}{10} & \TableICount{0}{10} & \TableICount{4}{10} & \TableICount{9}{10} & \TableICount{6}{10} & \TableICount{8}{10} & \TableICount{1}{10} & \TableICount{1}{10} & \TableICount{0}{10} & \TableICount{8}{10} \\
Needle Intrusion & III & \textbf{\TableICount{10}{10}} & \textbf{\TableICount{10}{10}} & \textbf{\TableICount{10}{10}} & \textbf{\TableICount{10}{10}} & \TableICount{0}{10} & \TableICount{9}{10} & \TableICount{0}{10} & \TableICount{1}{10} & \textbf{\TableICount{10}{10}} & \textbf{\TableICount{10}{10}} & \textbf{\TableICount{10}{10}} & \textbf{\TableICount{10}{10}} & \TableICount{1}{10} & \TableICount{0}{10} & \TableICount{0}{10} & \textbf{\TableICount{10}{10}} \\
\midrule
& & \multicolumn{16}{c}{\textbf{Shunt Insertion (Real)} --- ACT natural success: 14/20 (70\%)} \\
\rowcolor{black!9}
No failure & -- & \textbf{\TableICount{0}{20}} & \textbf{\TableICount{0}{20}} & \TableICount{1}{20} & \TableICount{1}{20} & \textbf{\TableICount{0}{20}} & \TableICount{2}{20} & \TableICount{1}{20} & \textbf{\TableICount{0}{20}} & \textbf{\TableICount{0}{20}} & \TableICount{1}{20} & \textbf{\TableICount{0}{20}} & \textbf{\TableICount{0}{20}} & \TableICount{0}{20} & \TableICount{0}{20} & \textbf{\TableICount{0}{20}} & \textbf{\TableICount{0}{20}} \\
\rowcolor{black!9}
Insertion Misalignment & II & \TableICount{6}{10} & \TableICount{8}{10} & \TableICount{6}{10} & \TableICount{9}{10} & \TableICount{0}{10} & \TableICount{0}{10} & \TableICount{0}{10} & \TableICount{0}{10} & \textbf{\TableICount{10}{10}} & \textbf{\TableICount{10}{10}} & \TableICount{4}{10} & \textbf{\TableICount{10}{10}} & \TableICount{0}{10} & \TableICount{0}{10} & \TableICount{5}{10} & \TableICount{0}{10} \\
Shunt Disturbance & III & \TableICount{5}{10} & \textbf{\TableICount{10}{10}} & \TableICount{2}{10} & \TableICount{6}{10} & \TableICount{0}{10} & \TableICount{0}{10} & \TableICount{1}{10} & \TableICount{0}{10} & \TableICount{7}{10} & \textbf{\TableICount{10}{10}} & \TableICount{3}{10} & \textbf{\TableICount{10}{10}} & \TableICount{10}{10} & \TableICount{9}{10} & \textbf{\TableICount{10}{10}} & \TableICount{5}{10} \\
Lateral Grasp Drift & IV & \TableICount{2}{10} & \textbf{\TableICount{10}{10}} & \TableICount{1}{10} & \textbf{\TableICount{10}{10}} & \TableICount{0}{10} & \TableICount{2}{10} & \TableICount{0}{10} & \TableICount{0}{10} & \TableICount{8}{10} & \textbf{\TableICount{10}{10}} & \TableICount{2}{10} & \TableICount{9}{10} & \TableICount{0}{10} & \TableICount{0}{10} & \TableICount{5}{10} & \TableICount{1}{10} \\
Camera Freeze & IV & \textbf{\TableICount{10}{10}} & \TableICount{9}{10} & \TableICount{2}{10} & \TableICount{2}{10} & \TableICount{3}{10} & \TableICount{2}{10} & \TableICount{0}{10} & \TableICount{0}{10} & \TableICount{0}{10} & \TableICount{1}{10} & \TableICount{0}{10} & \TableICount{0}{10} & \TableICount{0}{10} & \TableICount{0}{10} & \TableICount{0}{10} & \TableICount{0}{10} \\
\midrule
Overall FAR &  & \textbf{0.0\%} & 2.5\% & 2.5\% & 3.8\% & 1.2\% & 7.5\% & 1.2\% & \textbf{0.0\%} & 2.5\% & 2.5\% & 1.2\% & 2.5\% & 12.5\% & 16.2\% & \textbf{0.0\%} & 2.5\% \\
Overall FDR &  & 46.2\% & \textbf{95.6\%} & 46.9\% & 83.4\% & 23.4\% & 19.7\% & 1.6\% & 4.1\% & 47.2\% & 61.9\% & 25.0\% & 42.2\% & 20.3\% & 17.8\% & 29.4\% & 32.2\% \\
\midrule
\multicolumn{18}{@{}l@{}}{\textbf{(B) Calibrated time-varying bands: pooled FAR and FDR}} \\
Overall FAR &  & 5.0\% & \textbf{0.0\%} & 6.2\% & 3.8\% & 2.5\% & 6.2\% & \textbf{0.0\%} & 2.5\% & 1.2\% & 1.2\% & 1.2\% & 5.0\% & -- & -- & 17.5\% & 3.8\% \\
Overall FDR &  & 50.6\% & \textbf{87.5\%} & 42.5\% & 80.9\% & 31.2\% & 18.8\% & 2.5\% & 5.6\% & 53.4\% & 65.3\% & 27.5\% & 42.5\% & -- & -- & 10.0\% & 32.2\% \\
\midrule
\multicolumn{18}{@{}l@{}}{\textbf{(C) Retrospective thresholds: FDR at a 5\% \textcolor{black}{FAR budget}}} \\
Needle Pickup &  & 45.0\% & \textbf{99.2\%} & 67.5\% & 85.0\% & 35.8\% & 38.3\% & 1.7\% & 13.3\% & 59.2\% & 69.2\% & 51.7\% & 53.3\% & -- & -- & 54.2\% & 35.0\% \\
Ring Over Post &  & 30.8\% & \textbf{93.3\%} & 27.5\% & 83.3\% & 26.7\% & 10.8\% & 10.8\% & 0.8\% & 21.7\% & 66.7\% & 31.7\% & 44.2\% & -- & -- & 44.2\% & 25.0\% \\
Tissue Retraction &  & \textbf{100.0\%} & 97.5\% & \textbf{100.0\%} & \textbf{100.0\%} & 42.5\% & 75.0\% & 22.5\% & 12.5\% & 97.5\% & \textbf{100.0\%} & 95.0\% & 95.0\% & -- & -- & 80.0\% & 67.5\% \\
Shunt Insertion &  & 62.5\% & \textbf{100.0\%} & 30.0\% & 72.5\% & 20.0\% & 10.0\% & 2.5\% & 0.0\% & 82.5\% & 77.5\% & 45.0\% & 75.0\% & -- & -- & 57.5\% & 27.5\% \\
\midrule
Overall FDR &  & 48.8\% & \textbf{96.9\%} & 51.9\% & 84.7\% & 31.2\% & 29.1\% & 7.8\% & 6.9\% & 52.8\% & 73.1\% & 48.8\% & 57.8\% & -- & -- & 54.1\% & 34.4\% \\
\bottomrule
}
\sbox{\TableIBox}{%
\begin{tabular}{@{}lc*{16}{c}@{}}\TableIContent\end{tabular}%
}
\ifdim\wd\TableIBox>\textwidth
  \PackageError{table-i}{Natural width \the\wd\TableIBox\space exceeds \the\textwidth}{Reduce spacing; retain 8-point text.}
\fi
\begin{tabular*}{\textwidth}{@{\extracolsep{\fill}}lc*{16}{c}@{}}
\TableIContent
\end{tabular*}
\par\vspace{2pt}
\begin{minipage}{\textwidth}
\fontsize{8}{9.5}\selectfont\color{black}
Gray rows: no deliberate intervention. Categories: I grasp; II spatial; III scene; IV sensing/actuation.
\end{minipage}
\endgroup
\par\vspace{-4mm}
\end{table*}

\noindent\textbf{Failure modes and evaluation protocol.}
The evaluation contains 20 task-specific failure modes spanning four broad categories: (i) grasp acquisition and retention failures; (ii) spatial-configuration errors; (iii) scene disturbances; and (iv) sensing and actuation anomalies. The failure set comprises 90 naturally occurring ACT failures and 230 failures induced through changes to initial conditions, the scene, sensing, or actuation. Ring Drop and Tissue Grasp Loss are induced by gripper-opening overrides, Lateral Grasp Drift by translation biases, and Execution Mismatch by altered translation gains and cross-axis coupling. Needle and Ring Slippage use imposed object velocities. Other interventions alter initial object placement (OOD Initial Condition), restrain tissue (Fixed Tissue Target), displace tissue or the shunt (Tissue/Shunt Disturbance), or introduce an extraneous needle (Needle Intrusion). Blood Occlusion uses image-space blood-droplet overlays and blur on both views, while Camera Freeze repeats captured frames. Table~\ref{tab:failure_detection} provides per-mode detection results and natural policy success rates measured on separate unperturbed rollouts.

For each task, we use \(N=19\) successful policy rollouts for conformal calibration and set \(\alpha=0.05\) for all continuous-score detectors. The threshold is therefore the maximum of the 19 calibration episode maxima. The calibration rollouts are disjoint from the evaluation rollouts. For rollout-level evaluation, each simulated task includes 20 failed policy rollouts per failure mode and 20 held-out successful policy rollouts. Each real-robot task includes 10 failed policy rollouts per failure mode and 20 held-out successful policy rollouts. In total, the evaluation set contains 320 failed rollouts and 80 successful rollouts. We average three annotators' wall-clock onset times per view, except for simulated induced failures, which retain the original labels. A failed rollout is counted as detected if the score for the evaluated view exceeds its calibrated threshold for at least one endpoint at or after its annotated onset; a successful rollout is counted as a false alarm if its score exceeds the threshold at any point during the rollout. Pre-onset alarms are reported separately and do not contribute to FDR.

We benchmark detection offline on recorded policy rollouts and evaluate latency through online streaming replay. The offline benchmark scores every valid endpoint to compute FDR and FAR. Online replay follows recorded timing and recomputes decisions from the latest available window or frame. We measure latency from onset to completion of the first alarm using post-onset input, including computation.

\noindent\textbf{Baselines.}
WM-NC denotes the proposed nonconformity score. WM-PE uses the same world model but scores endpoint prediction error, with deterministic predictions obtained by integrating the forward ODE from a fixed zero base latent. WM-PU adapts Ward et al.'s uncertainty score~\cite{ward2026foundational} using the mean coordinate-wise standard deviation of 16 endpoint latents sampled from our flow-matching model, rather than directly predicted standard deviations. NoAct removes action conditioning from WM-NC. We further compare against logpZO, adapted from FAILDetect~\cite{xu2025can}, and random network distillation (RND), adapted from the observation branch of FIPER~\cite{romer2026failure}. Both are trained on expert demonstrations using ACT-native visual tokens; logpZO also uses proprioception. Action-chunk entropy (ACE) adapts FIPER's action branch through histogram entropy of sampled ACT chunk positions. Action ensemble uncertainty (AEU) adapts Thompson et al.~\cite{thompson2025early} using signed changes in positional disagreement among three ACT policies. Finally, our Sentinel-inspired VLM monitor~\cite{agia2025unpacking} uses zero-shot Qwen3-VL-8B-Instruct~\cite{bai2025qwen3vl} with task and failure descriptions and each view's observed video history sampled at 1 FPS, returning uncalibrated binary decisions online. All other detectors use the same calibration and evaluation rollout splits.

\begin{figure*}[!t]
    \centering
    \includegraphics[width=0.98\textwidth,trim={0pt 2pt 0pt 0pt},clip]{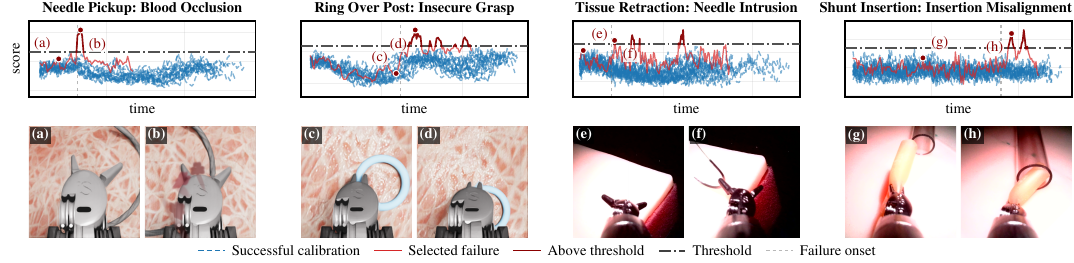}
    \caption{Representative wrist-view WM-NC score traces across four tasks. Frames (a)--(h) correspond to the labeled points on the traces.}
    \label{fig:failure_example}
\end{figure*}

\subsection{Experimental Results}

\noindent\textbf{Overall Performance.}
As shown in Table~\ref{tab:failure_detection}(A), wrist-view WM-NC achieves the highest overall FDR among the offline detectors at \(95.6\%\) with \(2.5\%\) FAR, compared with \(61.9\%\) FDR for logpZO at the same FAR. It also outperforms WM-PE and WM-PU using the same world model. We also assess the contribution of commanded actions by evaluating WM-NC without action conditioning (NoAct). Although NoAct retains strong detection performance, removing actions reduces wrist-view FDR by 12.2 percentage points to \(83.4\%\), indicating that actions provide useful additional context. To examine the effect of calibration, we evaluate time-varying conformal bands adapted from FAILDetect~\cite{xu2025can}, using frozen expert-derived templates for all continuous-score detectors. In Table~\ref{tab:failure_detection}(B), wrist-view WM-NC achieves \(87.5\%\) FDR with no false alarms on test successes. Under a retrospective \(5\%\) FAR budget per task, wrist-view WM-NC achieves \(96.9\%\) FDR in Table~\ref{tab:failure_detection}(C). Table~\ref{tab:wm_nc_categories} further compares WM-NC across failure categories and origins without recalibration, showing that spatial-configuration errors and natural failures remain more challenging for wrist-view detection. Fig.~\ref{fig:failure_example} shows representative wrist-view WM-NC score traces.

\begin{table}[!t]
\begingroup
\color{black}
\arrayrulecolor{black}
\fontsize{8}{9.5}\selectfont
\makeatletter
\renewcommand{\fnum@table}{\textcolor{black}{TABLE~\thetable}}
\makeatother
\caption{\textcolor{black}{WM-NC FDR by failure category and origin}}
\label{tab:wm_nc_categories}
\centering
\fontsize{8}{9.5}\selectfont
\setlength{\tabcolsep}{1.5pt}
\renewcommand{\arraystretch}{1.00}
\newsavebox{\TableIIBox}
\newcommand{\TableIIContent}{%
\toprule
& \multicolumn{4}{c}{\textbf{Category FDR}}
& \multicolumn{2}{c}{\textbf{Origin FDR}} \\
\cmidrule(lr){2-5}\cmidrule(l){6-7}
\textbf{View} & I & II & III & IV & Natural & Induced \\
\midrule
Fixed & 24.4\% & 18.0\% & 77.5\% & 55.0\% & 20.0\% & 56.5\% \\
Wrist & \textbf{96.7\%} & \textbf{82.0\%} & \textbf{98.8\%} & \textbf{99.0\%} & \textbf{87.8\%} & \textbf{98.7\%} \\
\bottomrule
}
\sbox{\TableIIBox}{%
\begin{tabular}{@{}l*{6}{c}@{}}\TableIIContent\end{tabular}%
}
\ifdim\wd\TableIIBox>\columnwidth
  \PackageError{table-ii}{Natural width \the\wd\TableIIBox\space exceeds \the\columnwidth}{Reduce spacing; retain 8-point text.}
\fi
\begin{tabular*}{\columnwidth}{@{\extracolsep{\fill}}l*{6}{c}@{}}
\TableIIContent
\end{tabular*}
\endgroup
\end{table}

\noindent\textbf{Detection Latency.}
Table~\ref{tab:onset_latency} reports detection latency measured by running each detector as recorded episodes are replayed. For each non-VLM detector, we replay failure episodes it detected in the evaluation reported in Table~\ref{tab:failure_detection}(A); for the VLM, we replay all test episodes. Panels (A)--(B) show that wrist-view WM-NC detects all selected failures by rollout end, with a mean delay of \(0.98\)~s. Panel (C) reports WM-NC wrist minus comparator delay (\(\Delta\)) on \(n\) shared on-time detections, using view-specific mean onsets. Wrist-view WM-NC has shorter mean onset-relative delays in all pairs except the fixed-view WM-NC comparison.

\begin{table}[!t]
\begingroup
\color{black}\arrayrulecolor{black}
\fontsize{8}{9.5}\selectfont
\makeatletter
\renewcommand{\fnum@table}{\textcolor{black}{TABLE~\thetable}}
\makeatother
\caption{\textcolor{black}{WM-NC wrist latency and paired comparisons}}
\label{tab:onset_latency}
\centering
\fontsize{8}{9.5}\selectfont
\setlength{\tabcolsep}{2.5pt}
\renewcommand{\arraystretch}{1.02}
\newcommand{\TableIIIAContent}{%
\toprule
\multicolumn{2}{@{}l@{}}{\textbf{(A) WM-NC Wrist Latency}} \\
\textbf{Failure mode} & \textbf{Mean $\boldsymbol{\pm}$ SD (s)} \\
\midrule
Needle Drop & 0.44\,$\pm$\,0.54 \\
Insecure Grasp (Needle) & 1.77\,$\pm$\,0.80 \\
Blood Occlusion (Needle) & 0.07\,$\pm$\,0.00 \\
OOD Initial Condition & 3.44\,$\pm$\,1.02 \\
Execution Mismatch & 1.06\,$\pm$\,0.44 \\
Needle Slippage & 0.44\,$\pm$\,0.26 \\
\midrule
Post Misalignment & 2.33\,$\pm$\,1.89 \\
Insecure Grasp (Ring) & 1.94\,$\pm$\,4.83 \\
Ring Drop & 0.32\,$\pm$\,0.41 \\
Blood Occlusion (Ring) & 0.07\,$\pm$\,0.00 \\
Lateral Grasp Drift (Ring) & 0.41\,$\pm$\,0.09 \\
Ring Slippage & 0.38\,$\pm$\,0.12 \\
\midrule
Tissue Grasp Loss & 0.07\,$\pm$\,0.01 \\
Fixed Tissue Target & 0.34\,$\pm$\,0.39 \\
Tissue Disturbance & 1.06\,$\pm$\,1.63 \\
Needle Intrusion & 1.18\,$\pm$\,1.75 \\
\midrule
Insertion Misalignment & 1.58\,$\pm$\,0.81 \\
Shunt Disturbance & 0.93\,$\pm$\,0.79 \\
Lateral Grasp Drift (Shunt) & 0.25\,$\pm$\,0.17 \\
Camera Freeze & 1.70\,$\pm$\,1.97 \\
\midrule
\textbf{Overall} & 0.98\,$\pm$\,1.71 \\
\bottomrule
}
\newcommand{\TableIIIBContent}{%
\toprule
\multicolumn{2}{@{}l@{}}{\textbf{(B) WM-NC Wrist}} \\
\multicolumn{2}{@{}l@{}}{\textbf{Alarm Summary}} \\
\midrule
Post-onset & 306/306 \\
Pre-onset & 15/306 (4.9\%) \\
\bottomrule
}
\newcommand{\TableIIICContent}{%
\toprule
\multicolumn{3}{@{}l@{}}{\textbf{(C) Paired Comparisons}} \\
\textbf{Method} & \(\boldsymbol{n}\) & \(\boldsymbol{\Delta}\) \textbf{(s)} \\
\midrule
WM-NC F & 146 & $+0.03$ \\
WM-PE F & 75 & $-0.31$ \\
WM-PE W & 62 & $-2.04$ \\
WM-PU F & 2 & $-0.32$ \\
WM-PU W & 5 & $-3.45$ \\
logpZO F & 146 & $-0.34$ \\
logpZO W & 192 & $-0.54$ \\
RND F & 79 & $-0.87$ \\
RND W & 129 & $-0.94$ \\
VLM F & 49 & $-1.89$ \\
VLM W & 35 & $-2.28$ \\
ACE & 91 & $-0.95$ \\
AEU & 100 & $-1.39$ \\
\bottomrule
}
\newsavebox{\TableIIIALeftBox}
\newsavebox{\TableIIIBSummaryBox}
\newsavebox{\TableIIICRightBox}
\newsavebox{\TableIIIBox}
\newlength{\TableIIILeftWidth}
\newlength{\TableIIIRightWidth}
\sbox{\TableIIIALeftBox}{\begin{tabular}[t]{@{}lr@{}}\TableIIIAContent\end{tabular}}
\sbox{\TableIIIBSummaryBox}{\setlength{\tabcolsep}{2pt}\begin{tabular}[t]{@{}lr@{}}\TableIIIBContent\end{tabular}}
\sbox{\TableIIICRightBox}{\begin{tabular}[t]{@{}lrr@{}}\TableIIICContent\end{tabular}}
\setlength{\TableIIILeftWidth}{\wd\TableIIIALeftBox}
\setlength{\TableIIIRightWidth}{\dimexpr\columnwidth-\TableIIILeftWidth-8pt\relax}
\ifdim\wd\TableIIIBSummaryBox>\TableIIIRightWidth
  \PackageError{review-tables}{Table III replay summary exceeds its panel width}{Retain 8-point text; revise the summary labels.}
\fi
\ifdim\wd\TableIIICRightBox>\TableIIIRightWidth
  \PackageError{review-tables}{Table III right natural width \the\wd\TableIIICRightBox\space exceeds available \the\TableIIIRightWidth}{Retain 8-point text; adjust panel spacing.}
\fi
\sbox{\TableIIIBox}{%
\begin{tabular*}{\TableIIILeftWidth}[t]{@{\extracolsep{\fill}}lr@{}}\TableIIIAContent\end{tabular*}%
\hspace{8pt}%
\begin{minipage}[t]{\TableIIIRightWidth}
{\setlength{\tabcolsep}{2pt}\begin{tabular*}{\linewidth}[t]{@{\extracolsep{\fill}}lr@{}}\TableIIIBContent\end{tabular*}}
\par\vspace{10pt}
\begin{tabular*}{\linewidth}[t]{@{\extracolsep{\fill}}lrr@{}}\TableIIICContent\end{tabular*}
\par\vspace{4pt}
\fontsize{8}{9.5}\selectfont\color{black}\raggedright
\noindent F/W: fixed/wrist.\par\vspace{3pt}
\mbox{\(n\): shared detections.}\par\vspace{3pt}
\(\Delta\): mean delay difference.\par
\end{minipage}%
}
\ifdim\wd\TableIIIBox>\columnwidth
  \PackageError{review-tables}{Table III natural width \the\wd\TableIIIBox\space exceeds \the\columnwidth}{Retain 8-point text; tighten the layout.}
\fi
\ifdim\dimexpr\ht\TableIIIBox+\dp\TableIIIBox+70pt\relax>\textheight
  \PackageError{review-tables}{Table III exceeds the page-height allowance}{Retain 8-point text; revise layout.}
\fi
\usebox{\TableIIIBox}
\endgroup
\end{table}

\noindent\textbf{Effect of Camera View.}
Since clinical deployment may require different camera placements or endoscopic views, we report view-specific detectors to isolate the effect of camera geometry. Table~\ref{tab:failure_detection} shows that the wrist view generally provides stronger WM-NC detection than the fixed view. This pattern is consistent with manipulation-related failure cues being localized around the instrument tip, where the wrist camera provides a closer view of tool--object interactions. Fixed-view performance remains limited for several manipulation failures and OOD initial conditions, with some baselines achieving higher FDR.

We also evaluate camera fusion and pairing WM-NC with the marginal observation score logpZO (Table~\ref{tab:wm_nc_fusion}). We combine episode-level alarms using OR/AND on common eligible endpoints at the thresholds in Table~\ref{tab:failure_detection}(A), and train a single-branch world model on channel-wise concatenated fixed- and wrist-view VAE latents (Joint). Camera fusion offers no substantial FDR gain over wrist-only WM-NC, whereas OR fusion with logpZO improves fixed-view FDR without increasing the observed FAR.

\begin{table}[!t]
\begingroup
\color{black}\arrayrulecolor{black}
\fontsize{8}{9.5}\selectfont
\makeatletter
\renewcommand{\fnum@table}{\textcolor{black}{TABLE~\thetable}}
\makeatother
\caption{\textcolor{black}{Camera-view fusion and combination with logpZO}}
\label{tab:wm_nc_fusion}
\centering
\fontsize{8}{9.5}\selectfont
\setlength{\tabcolsep}{2pt}
\renewcommand{\arraystretch}{1.00}
\newsavebox{\TableVBox}
\newcommand{\TableVContent}{%
\toprule
 & \multicolumn{3}{c}{\textbf{WM-NC: both views}} & \multicolumn{2}{c}{\textbf{WM-NC OR logpZO}} \\
\cmidrule(lr){2-4}\cmidrule(l){5-6}
\textbf{Metric} & OR & AND & Joint & Fixed & Wrist \\
\midrule
\textbf{FDR} & 96.2\% & 45.6\% & 73.1\% & 64.1\% & \textbf{97.2\%} \\
\textbf{FAR} & 2.5\% & \textbf{0.0\%} & 1.2\% & \textbf{0.0\%} & 3.8\% \\
\bottomrule
}
\sbox{\TableVBox}{\begin{tabular}{@{}l*{5}{c}@{}}\TableVContent\end{tabular}}
\ifdim\wd\TableVBox>\columnwidth
  \PackageError{review-tables}{TableV natural width \the\wd\TableVBox\space exceeds \the\columnwidth}{Retain 8-point text; revise columns.}
\fi
\begin{tabular*}{\columnwidth}{@{\extracolsep{\fill}}l*{5}{c}@{}}
\TableVContent
\end{tabular*}
\endgroup
\end{table}

\noindent\textbf{Expert--Policy Score Comparison.}
We compare episode-maximum WM-NC scores between expert-validation demonstrations and successful policy rollouts. Across all tasks and views, policy-rollout medians are \(2.2\)--\(8.4\%\) lower and interquartile ranges \(2.2\)--\(60.9\%\) narrower than the expert values, supporting calibration on successful policy rollouts.

\section{Sensitivity and Robustness}
\subsection{Parameter and Data Sensitivity}

\begin{figure}[!t]
\begingroup
\makeatletter
\renewcommand{\fnum@figure}{Fig.~\thefigure}
\makeatother
\centering
\includegraphics[width=\columnwidth,trim={0pt 6pt 0pt 0pt},clip]{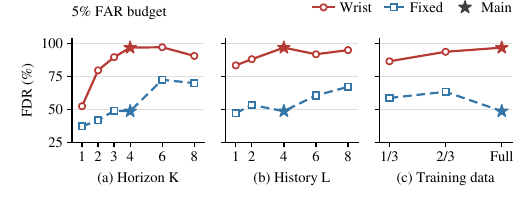}
\caption{WM-NC sensitivity at a retrospective \(5\%\) FAR budget.}
\label{fig:ablation_sensitivity}
\endgroup
\end{figure}

We also vary the endpoint horizon \(K\), history length \(L\), and world-model training-data size independently in Fig.~\ref{fig:ablation_sensitivity}. Over the tested ranges, FDR is most sensitive to \(K\): very short horizons limit detection, consistent with deviations becoming more apparent over a longer window. The benefits of additional history and training data vary by view: the main setting performs well in the wrist view, but trails several variants in the fixed view. We also evaluate online wrist-view latency across the \(K\) and \(L\) sweeps (Table~\ref{tab:kl_latency}). Delay increases with \(L\) but not consistently with \(K\): longer windows require more initial observations, while post-onset latency also depends on how quickly failure scores exceed the calibrated threshold.

\begin{table}[!t]
\begingroup
\color{black}\arrayrulecolor{black}
\makeatletter
\renewcommand{\fnum@table}{\textcolor{black}{TABLE~\thetable}}
\makeatother
\caption{\textcolor{black}{Wrist-view WM-NC latency (s) across \(K\) and \(L\)}}
\label{tab:kl_latency}
\centering
\fontsize{8}{9.5}\selectfont
\setlength{\tabcolsep}{2pt}
\renewcommand{\arraystretch}{1.00}
\begin{tabular*}{\columnwidth}{@{\extracolsep{\fill}}l*{5}{c}@{}}
\toprule
\textbf{Sweep} & \textbf{1} & \textbf{2} & \textbf{4} & \textbf{6} & \textbf{8} \\
\midrule
\(K\) (\(L=4\)) & 0.86 & 0.53 & \textbf{0.72} & 0.62 & 0.93 \\
\(L\) (\(K=4\)) & 0.63 & 0.68 & \textbf{0.72} & 0.74 & 0.84 \\
\bottomrule
\end{tabular*}
\par\smallskip
\begin{minipage}{\columnwidth}
\fontsize{7}{8}\selectfont
Mean onset-relative delay (s) on 145 wrist-view episodes detected before rollout end by all evaluated configurations. Bold: main setting (\(K=L=4\)).
\end{minipage}
\endgroup
\end{table}

\subsection{Uncertainty and Robustness}

Beyond parameter sensitivity, we assess statistical uncertainty and variation across calibration splits and training seeds. Table~\ref{tab:wm_nc_robustness} reports \(95\%\) Wilson confidence intervals (CI) conditional on the main model's thresholds. We also evaluate 500 random 19/20 calibration/test splits of successful rollouts with the model and failure episodes fixed, and four training seeds (0--3; seed 0 is the main model) on the same evaluation episodes. Both analyses report mean and sample SD. Wrist-view FDR remains high across calibration splits and training seeds. Relative to the four-seed average, the main model has higher wrist-view FDR and lower fixed-view FDR.

\begin{table}[!t]
\begingroup
\color{black}\arrayrulecolor{black}
\fontsize{8}{9.5}\selectfont
\makeatletter
\renewcommand{\fnum@table}{\textcolor{black}{TABLE~\thetable}}
\makeatother
\caption{\textcolor{black}{WM-NC uncertainty and robustness}}
\label{tab:wm_nc_robustness}
\centering
\fontsize{8}{9.5}\selectfont
\setlength{\tabcolsep}{1.5pt}
\renewcommand{\arraystretch}{1.00}
\newsavebox{\RobustnessTableBox}
\newcommand{\RobustnessTableContent}{%
\toprule
\textbf{Analysis} & \textbf{View} & \textbf{FAR} & \textbf{FDR} & \textbf{FDR at 5\% FAR budget} \\
\midrule
\textbf{(A) Main} & Fixed & [0.0, 4.6]\% & [40.9, 51.7]\% & [43.3, 54.2]\% \\
95\% CI & Wrist & [0.7, 8.7]\% & [92.8, 97.4]\% & [94.3, 98.3]\% \\
\midrule
\textbf{(B) 500 splits} & Fixed & 5.0\,$\pm$\,3.3\% & 47.7\,$\pm$\,1.4\% & 50.0\,$\pm$\,2.0\% \\
Mean $\pm$ SD & Wrist & 5.2\,$\pm$\,3.5\% & 94.9\,$\pm$\,1.8\% & 97.4\,$\pm$\,0.6\% \\
\midrule
\textbf{(C) 4 seeds} & Fixed & 3.1\,$\pm$\,3.0\% & 48.8\,$\pm$\,2.3\% & 54.8\,$\pm$\,8.7\% \\
Mean $\pm$ SD & Wrist & 4.7\,$\pm$\,4.1\% & 93.4\,$\pm$\,2.4\% & 94.6\,$\pm$\,2.4\% \\
\bottomrule
}
\sbox{\RobustnessTableBox}{\begin{tabular}{@{}llccc@{}}\RobustnessTableContent\end{tabular}}
\ifdim\wd\RobustnessTableBox>\columnwidth
  \PackageError{review-tables}{Robustness table natural width \the\wd\RobustnessTableBox\space exceeds \the\columnwidth}{Retain 8-point text; revise columns.}
\fi
\begin{tabular*}{\columnwidth}{@{\extracolsep{\fill}}llccc@{}}
\RobustnessTableContent
\end{tabular*}
\endgroup
\end{table}

\section{Discussion and Conclusion}

We presented FoMo-FD, a robot failure detection framework built on a flow-matching world model and evaluated on surgically relevant robot manipulation tasks. In particular, the gap between WM-NC and WM-PE/WM-PU under the same world model highlights the importance of scoring. WM-PE measures error against one predicted endpoint and may penalize valid variability, while WM-PU measures predictive uncertainty without assessing the observed endpoint. WM-NC instead scores that endpoint under the learned conditional transport. However, an already abnormal history may weaken this conditional score, potentially contributing to the fixed-view OOD limitation and motivating combination with marginal observation scores such as logpZO. The wrist-view comparisons with the evaluated observation-based, action-based, and VLM baselines suggest value in short-window visual dynamics with action context, although NoAct retains substantial detection ability. The fixed-view comparisons, however, show a less consistent advantage: at the same FAR budget, the main configuration trails some baselines, whereas variants using longer windows or smaller training subsets surpass them in our sensitivity study. The need for learned monitoring also depends on the failure type: dedicated checks may suffice for simple sensor faults such as camera freeze.

Another limitation is our reliance on single-operator demonstrations and synthetic tissue models. Extending evaluation to ex vivo or in vivo settings calls for larger, more diverse datasets capturing variation in tissue appearance and deformation, operator behavior, and camera conditions. Longer tasks such as suturing also require coverage of different task phases and their transitions. Greater model capacity may help represent this diversity, while task-phase context or longer-term memory may help capture dependencies beyond the local observation window. The same monitoring formulation could be applied to both settings, although its effectiveness there remains to be validated.

Extending the method to these settings would still require a small set of successful policy rollouts for calibration, with false-alarm guarantees depending on exchangeability with future successful rollouts. The current framework also does not address failure recovery. Inspired by prior work coupling anomaly detection with language-model reasoning~\cite{sinha2024realtime}, future work will explore integrating local monitoring with procedural checks and failure-category identification to support targeted recovery.

\bibliographystyle{IEEEtran}
\bibliography{refs_short}

@IEEEtranBSTCTL{IEEE:BSTcontrol,
  CTLuse_forced_etal = "yes",
  CTLmax_names_forced_etal = "6",
  CTLnames_show_etal = "1",
  CTLdash_repeated_names = "no",
}

@inproceedings{kawaharazuka2024robotic,
  title={Robotic constrained imitation learning for the peg transfer task in fundamentals of laparoscopic surgery},
  author={Kawaharazuka, Kento and Okada, Kei and Inaba, Masayuki},
  booktitle={Proc. IEEE Int. Conf. Robot. Autom. (ICRA)},
  pages={606--612},
  year={2024},
  organization={IEEE}
}

@inproceedings{kim2025surgical,
  title={Surgical Robot Transformer ({SRT}): Imitation Learning for Surgical Tasks},
  author={Kim, Ji Woong and Zhao, Tony Z and Schmidgall, Samuel and Deguet, Anton and Kobilarov, Marin and Finn, Chelsea and Krieger, Axel},
  booktitle={Conf. Robot Learn. (CoRL)},
  pages={130--144},
  year={2025},
  organization={PMLR}
}

@article{kim2025srt,
  title={{SRT-H}: A hierarchical framework for autonomous surgery via language-conditioned imitation learning},
  author={Kim, Ji Woong and Chen, Juo-Tung and Hansen, Pascal and Shi, Lucy Xiaoyang and Goldenberg, Antony and Schmidgall, Samuel and Scheikl, Paul Maria and Deguet, Anton and White, Brandon M and Tsai, De Ru and others},
  journal={Sci. Robot.},
  volume={10},
  number={104},
  pages={eadt5254},
  year={2025},
  publisher={American Association for the Advancement of Science}
}

@article{haworth2026suturebot,
  title={{SutureBot}: A precision framework \& benchmark for autonomous end-to-end suturing},
  author={Haworth, Jesse and Chen, Juo-Tung and Nelson, Nigel and Kim, Ji Woong and Moghani, Masoud and Finn, Chelsea and Krieger, Axel},
  journal={Adv. Neural Inf. Process. Syst.},
  volume={38},
  year={2026}
}

@inproceedings{yasar2020real,
  title={Real-time context-aware detection of unsafe events in robot-assisted surgery},
  author={Yasar, Mohammad Samin and Alemzadeh, Homa},
  booktitle={Proc. IEEE/IFIP Int. Conf. Dependable Syst. Netw. (DSN)},
  pages={385--397},
  year={2020},
  organization={IEEE}
}

@inproceedings{kam2021confidence,
  title={A confidence-based supervised-autonomous control strategy for robotic vaginal cuff closure},
  author={Kam, Michael and Saeidi, Hamed and Hsieh, Michael H and Kang, Jin U and Krieger, Axel},
  booktitle={Proc. IEEE Int. Conf. Robot. Autom. (ICRA)},
  pages={12261--12267},
  year={2021},
  organization={IEEE}
}

@article{thompson2025early,
  title={Early failure detection in autonomous surgical soft-tissue manipulation via uncertainty quantification},
  author={Thompson, Jordan and Koe, Ronald and Le, Anthony and Goodman, Gabriella and Brown, Daniel S and Kuntz, Alan},
  journal={arXiv preprint arXiv:2501.10561},
  year={2025}
}

@article{empleo2026safe,
  title={Safe Uncertainty-Aware Learning Framework for Robotic Suturing},
  author={Empleo, Wilbert Peter and Kim, Yitaek and Kim, Hansoul and Savarimuthu, Thiusius Rajeeth and Iturrate, I{\~n}igo},
  journal={IEEE Trans. Med. Robot. Bionics},
  volume={8},
  number={1},
  pages={41--53},
  year={2026},
  publisher={IEEE}
}

@inproceedings{agia2025unpacking,
  title={Unpacking Failure Modes of Generative Policies: Runtime Monitoring of Consistency and Progress},
  author={Agia, Christopher and Sinha, Rohan and Yang, Jingyun and Cao, Ziang and Antonova, Rika and Pavone, Marco and Bohg, Jeannette},
  booktitle={Conf. Robot Learn. (CoRL)},
  pages={689--723},
  year={2025},
  organization={PMLR}
}

@inproceedings{zhou2025code,
  title={{Code-as-Monitor}: Constraint-aware visual programming for reactive and proactive robotic failure detection},
  author={Zhou, Enshen and Su, Qi and Chi, Cheng and Zhang, Zhizheng and Wang, Zhongyuan and Huang, Tiejun and Sheng, Lu and Wang, He},
  booktitle={Proc. IEEE/CVF Conf. Comput. Vis. Pattern Recognit. (CVPR)},
  pages={6919--6929},
  year={2025}
}

@article{gu2026safe,
  title={{SAFE}: Multitask failure detection for vision-language-action models},
  author={Gu, Qiao and Ju, Yuanliang and Sun, Shengxiang and Gilitschenski, Igor and Nishimura, Haruki and Itkina, Masha and Shkurti, Florian},
  journal={Adv. Neural Inf. Process. Syst.},
  volume={38},
  pages={40041--40076},
  year={2026}
}

@article{romer2026failure,
  title={Failure prediction at runtime for generative robot policies},
  author={R{\"o}mer, Ralf and Kobras, Adrian and Worbis, Luca and Schoellig, Angela},
  journal={Adv. Neural Inf. Process. Syst.},
  volume={38},
  pages={7631--7670},
  year={2026}
}

@INPROCEEDINGS{xu2025can, 
    AUTHOR    = {Chen Xu AND Tony Khuong Nguyen AND Emma Dixon AND Christopher Rodriguez AND Patrick Miller AND Robert Lee AND Paarth Shah AND Rares Andrei Ambrus AND Haruki Nishimura AND Masha Itkina}, 
    TITLE     = {Can We Detect Failures Without Failure Data? {Uncertainty}-Aware Runtime Failure Detection for Imitation Learning Policies}, 
    BOOKTITLE = {Robot.: Sci. Syst. (RSS)}, 
    YEAR      = {2025}, 
    ADDRESS   = {Los Angeles, CA, USA}, 
    MONTH     = {June}, 
    DOI       = {10.15607/RSS.2025.XXI.073} 
}

@inproceedings{zhou2026rc,
  title={{RC-NF}: Robot-Conditioned Normalizing Flow for Real-Time Anomaly Detection in Robotic Manipulation},
  author={Zhou, Shijie and Zhu, Bin and Yang, Jiarui and Zhao, Xiangyu and Chen, Jingjing and Jiang, Yu-Gang},
  booktitle={Proc. IEEE/CVF Conf. Comput. Vis. Pattern Recognit. (CVPR)},
  pages={43050--43060},
  year={2026}
}

@article{ward2026foundational,
  title={Foundational World Models Accurately Detect Bimanual Manipulator Failures},
  author={Ward, Isaac R and Ho, Michelle and Liu, Houjun and Feldman, Aaron and Vincent, Joseph and Kruse, Liam and Cheong, Sean and Eddy, Duncan and Kochenderfer, Mykel J and Schwager, Mac},
  journal={arXiv preprint arXiv:2603.06987},
  year={2026}
}

@article{angelopoulos2023conformal,
  title={Conformal prediction: A gentle introduction},
  author={Angelopoulos, Anastasios N and Bates, Stephen},
  journal={Found. Trends Mach. Learn.},
  volume={16},
  number={4},
  pages={494--591},
  year={2023},
  publisher={Emerald Publishing Limited}
}

@inproceedings{kazanzides2014open,
  title={An open-source research kit for the {da Vinci}{\textregistered} Surgical System},
  author={Kazanzides, Peter and Chen, Zihan and Deguet, Anton and Fischer, Gregory S and Taylor, Russell H and DiMaio, Simon P},
  booktitle={Proc. IEEE Int. Conf. Robot. Autom. (ICRA)},
  pages={6434--6439},
  year={2014},
  organization={IEEE}
}

@inproceedings{lin2024world,
  title     = {World Models for General Surgical Grasping},
  author    = {Lin, Hongbin and Li, Bin and Wong, Chun Wai and Rojas, Juan and Chu, Xiangyu and Au, Kwok Wai Samuel},
  booktitle = {Robot.: Sci. Syst. (RSS)},
  year      = {2024},
  address   = {Delft, Netherlands},
  month     = jul,
  doi       = {10.15607/RSS.2024.XX.041}
}

@inproceedings{koju2025surgical,
  title={Surgical vision world model},
  author={Koju, Saurabh and Bastola, Saurav and Shrestha, Prashant and Amgain, Sanskar and Shrestha, Yash Raj and Poudel, Rudra PK and Bhattarai, Binod},
  booktitle={MICCAI Workshop Data Eng. Med. Imag.},
  pages={1--10},
  year={2025},
  organization={Springer}
}

@misc{he2025cosmoshsurgical,
  title         = {{Cosmos-H-Surgical}: Learning Surgical Robot Policies from Videos via World Modeling},
  author        = {He, Yufan and Guo, Pengfei and Xu, Mengya and Li, Zhaoshuo and Myronenko, Andriy and Imans, Dillan and Liu, Bingjie and Yang, Dongren and Gu, Mingxue and Ji, Yongnan and Jin, Yueming and Zhao, Ren and Shen, Baiyong and Xu, Daguang},
  year          = {2025},
  eprint        = {2512.23162},
  archivePrefix = {arXiv},
  primaryClass  = {cs.RO},
  doi           = {10.48550/arXiv.2512.23162},
  note          = {arXiv:2512.23162}
}

@misc{turkcan2025suturing,
      title={Towards Suturing World Models: Learning Predictive Models for Robotic Surgical Tasks}, 
      author={Mehmet Kerem Turkcan and Mattia Ballo and Filippo Filicori and Zoran Kostic},
      year={2025},
      eprint={2503.12531},
      archivePrefix={arXiv},
      primaryClass={cs.CV},
      url={https://arxiv.org/abs/2503.12531}, 
}

@article{zbinden2026cosmos,
  title={{Cosmos-Surg-dVRK}: world foundation model-based automated online evaluation of surgical robot policy learning},
  author={Zbinden, Lukas and Nelson, Nigel and Chen, Juo-Tung and Chen, Xinhao and Kim, Ji Woong and Azizian, Mahdi and Krieger, Axel and Huver, Sean},
  journal={IEEE Robot. Autom. Lett.},
  year={2026},
  publisher={IEEE}
}

@misc{chen2025how,
      title={How Far Are Surgeons from Surgical World Models? {A} Pilot Study on Zero-shot Surgical Video Generation with Expert Assessment}, 
      author={Zhen Chen and Qing Xu and Jinlin Wu and Biao Yang and Yuhao Zhai and Geng Guo and Jing Zhang and Yinlu Ding and Nassir Navab and Jiebo Luo},
      year={2025},
      eprint={2511.01775},
      archivePrefix={arXiv},
      primaryClass={cs.CV},
      url={https://arxiv.org/abs/2511.01775}, 
}

@article{oquab2024dinov2,
  title={{DINOv2}: Learning Robust Visual Features without Supervision},
  author={Oquab, Maxime and Darcet, Timoth{\'e}e and Moutakanni, Th{\'e}o and Vo, Huy and Szafraniec, Marc and Khalidov, Vasil and Fernandez, Pierre and Haziza, Daniel and Massa, Francisco and El-Nouby, Alaaeldin and others},
  journal={Trans. Mach. Learn. Res.},
  year={2024}
}

@inproceedings{zhou2025dino,
  title={{DINO-WM}: World Models on Pre-trained Visual Features enable Zero-shot Planning},
  author={Zhou, Gaoyue and Pan, Hengkai and LeCun, Yann and Pinto, Lerrel},
  booktitle={Int. Conf. Mach. Learn. (ICML)},
  pages={79115--79135},
  year={2025},
  organization={PMLR}
}

@misc{baldassarre2025features,
      title={Back to the Features: {DINO} as a Foundation for Video World Models}, 
      author={Federico Baldassarre and Marc Szafraniec and Basile Terver and Vasil Khalidov and Francisco Massa and Yann LeCun and Patrick Labatut and Maximilian Seitzer and Piotr Bojanowski},
      year={2025},
      eprint={2507.19468},
      archivePrefix={arXiv},
      primaryClass={cs.CV},
      url={https://arxiv.org/abs/2507.19468}, 
}

@article{karypidis2026dino,
  title={{DINO-Foresight}: Looking into the future with {DINO}},
  author={Karypidis, Efstathios and Kakogeorgiou, Ioannis and Gidaris, Spyridon and Komodakis, Nikos},
  journal={Adv. Neural Inf. Process. Syst.},
  volume={38},
  pages={163779--163811},
  year={2026}
}

@misc{boduljak2025vfmf,
      title={{VFMF}: World Modeling by Forecasting Vision Foundation Model Features}, 
      author={Gabrijel Boduljak and Yushi Lan and Christian Rupprecht and Andrea Vedaldi},
      year={2025},
      eprint={2512.11225},
      archivePrefix={arXiv},
      primaryClass={cs.CV},
      url={https://arxiv.org/abs/2512.11225}, 
}

@inproceedings{lipman2023flow,
  title={Flow Matching for Generative Modeling},
  author={Lipman, Yaron and Chen, Ricky T. Q. and Ben-Hamu, Heli and Nickel, Maximilian and Le, Matthew},
  booktitle={Int. Conf. Learn. Represent. (ICLR)},
  year={2023}
}

@inproceedings{moghani2025sufia,
  title={{SuFIA-BC}: Generating high quality demonstration data for visuomotor policy learning in surgical subtasks},
  author={Moghani, Masoud and Nelson, Nigel and Ghanem, Mohamed and Diaz-Pinto, Andres and Hari, Kush and Azizian, Mahdi and Goldberg, Ken and Huver, Sean and Garg, Animesh},
  booktitle={Proc. IEEE Int. Conf. Robot. Autom. (ICRA)},
  pages={4534--4541},
  year={2025},
  organization={IEEE}
}

@INPROCEEDINGS{zhao2023learning, 
    AUTHOR    = {Tony Z. Zhao AND Vikash Kumar AND Sergey Levine AND Chelsea Finn}, 
    TITLE     = {Learning Fine-Grained Bimanual Manipulation with Low-Cost Hardware}, 
    BOOKTITLE = {Robot.: Sci. Syst. (RSS)}, 
    YEAR      = {2023}, 
    ADDRESS   = {Daegu, Republic of Korea}, 
    MONTH     = {July}, 
    DOI       = {10.15607/RSS.2023.XIX.016} 
}

@article{bai2025qwen3vl,
  title={{Qwen3-VL} Technical Report},
  author={Bai, Shuai and Cai, Yuxuan and Chen, Ruizhe and Chen, Keqin and Chen, Xionghui and Cheng, Zesen and Deng, Lianghao and Ding, Wei and Gao, Chang and Ge, Chunjiang and others},
  journal={arXiv preprint arXiv:2511.21631},
  year={2025}
}

@article{hutchinson2022analysis,
  title={Analysis of executional and procedural errors in dry-lab robotic surgery experiments},
  author={Hutchinson, Kay and Li, Zongyu and Cantrell, Leigh A. and Schenkman, Noah S. and Alemzadeh, Homa},
  journal={Int. J. Med. Robot. Comput. Assist. Surg.},
  volume={18},
  number={3},
  pages={e2375},
  year={2022},
  doi={10.1002/rcs.2375}
}

@inproceedings{li2022runtime,
  title={Runtime Detection of Executional Errors in Robot-Assisted Surgery},
  author={Li, Zongyu and Hutchinson, Kay and Alemzadeh, Homa},
  booktitle={Proc. IEEE Int. Conf. Robot. Autom. (ICRA)},
  year={2022},
  organization={IEEE}
}

@article{shao2024think,
  title={Think Step by Step: Chain-of-Gesture Prompting for Error Detection in Robotic Surgical Videos},
  author={Shao, Zhimin and Xu, Jialang and Stoyanov, Danail and Mazomenos, Evangelos B. and Jin, Yueming},
  journal={IEEE Robot. Autom. Lett.},
  volume={9},
  number={12},
  pages={11513--11520},
  year={2024},
  doi={10.1109/LRA.2024.3495452}
}

@article{roodabeh2026realtime,
  title={Real-Time Multimodal Activity-Aware Error Detection in Robot-Assisted Surgery},
  author={Roodabeh, Seyed Hamid Reza and Li, Zongyu and Alemzadeh, Homa},
  journal={arXiv preprint arXiv:2606.23593},
  year={2026}
}

@article{low2025cares,
  title={{CARES}: Collaborative Agentic Reasoning for Error Detection in Surgery},
  author={Low, Chang Han and Zhuo, Zhu and Wang, Ziyue and Xu, Jialang and Liu, Haofeng and Sirajudeen, Nazir and Boal, Matthew and Edwards, Philip J. and Stoyanov, Danail and Francis, Nader and Zhong, Jiehui and Gu, Di and Mazomenos, Evangelos B. and Jin, Yueming},
  journal={arXiv preprint arXiv:2508.08764},
  year={2025}
}

@inproceedings{gudovskiy2022cflow,
  title={{CFLOW-AD}: Real-Time Unsupervised Anomaly Detection with Localization via Conditional Normalizing Flows},
  author={Gudovskiy, Denis and Ishizaka, Shun and Kozuka, Kazuki},
  booktitle={Proc. IEEE/CVF Winter Conf. Appl. Comput. Vis. (WACV)},
  pages={98--107},
  year={2022}
}

@article{luo2024sample,
  title={Sample-efficient safety assurances using conformal prediction},
  author={Luo, Rachel and Zhao, Shengjia and Kuck, Jonathan and Ivanovic, Boris and Savarese, Silvio and Schmerling, Edward and Pavone, Marco},
  journal={Int. J. Robot. Res.},
  volume={43},
  number={9},
  pages={1409--1424},
  year={2024},
  doi={10.1177/02783649231221580}
}

@inproceedings{lindemann2023conformal,
  title={Conformal Prediction for {STL} Runtime Verification},
  author={Lindemann, Lars and Qin, Xin and Deshmukh, Jyotirmoy V. and Pappas, George J.},
  booktitle={Proc. ACM/IEEE Int. Conf. Cyber-Phys. Syst. (ICCPS)},
  pages={142--153},
  year={2023},
  doi={10.1145/3576841.3585927}
}

@inproceedings{sinha2024realtime,
  title={Real-Time Anomaly Detection and Reactive Planning with Large Language Models},
  author={Sinha, Rohan and Elhafsi, Amine and Agia, Christopher and Foutter, Matt and Schmerling, Edward and Pavone, Marco},
  booktitle={Robot.: Sci. Syst. (RSS)},
  year={2024},
  doi={10.15607/RSS.2024.XX.114}
}

@article{hwang2020efficiently,
  title={Efficiently Calibrating Cable-Driven Surgical Robots With {RGBD} Fiducial Sensing and Recurrent Neural Networks},
  author={Hwang, Minho and Thananjeyan, Brijen and Paradis, Samuel and Seita, Daniel and Ichnowski, Jeffrey and Fer, Danyal and Low, Thomas and Goldberg, Ken},
  journal={IEEE Robot. Autom. Lett.},
  volume={5},
  pages={5937--5944},
  year={2020},
  doi={10.1109/LRA.2020.3010746}
}

\end{document}